\def\year{2020}\relax
\documentclass[letterpaper]{article} 
\usepackage{aaai20}  
\usepackage{times}  
\usepackage{helvet} 
\usepackage{courier}  
\usepackage[hyphens]{url}  
\usepackage{graphicx} 
\usepackage{graphicx}  
\usepackage{multirow}
\usepackage{amsmath}
\usepackage{bm}
\usepackage{amssymb}
\title{Cascading Convolutional Color Constancy}
\author{
	Huanglin Yu,\textsuperscript{\rm 1}
	Ke Chen,\textsuperscript{\rm 1,\thanks{Corresponding author.}}
	Kaiqi Wang,\textsuperscript{\rm 1}
	Yanlin Qian,\textsuperscript{\rm 2}
	Zhaoxiang Zhang,\textsuperscript{\rm 3}
	Kui Jia\textsuperscript{\rm 1}\\
	\textsuperscript{\rm 1}South China University of Technology,
	\textsuperscript{\rm 2}Tampere University,
	\textsuperscript{\rm 3}Chinese Academy of Sciences\\
	eeyu.huanglin@mail.scut.edu.cn, chenk@scut.edu.cn, mswkq@mail.scut.edu.cn \\
	yanlin.qian@tuni.fi, zhaoxiang.zhang@ia.ac.cn, kuijia@scut.edu.cn\\
}
\begin{document}
	
	\maketitle
	
	\begin{abstract}
		Regressing the illumination of a scene from the representations of object appearances is popularly adopted in computational color constancy.
		However, it's still challenging due to intrinsic appearance and label ambiguities caused by unknown illuminants, diverse reflection property of materials and extrinsic imaging factors (such as different camera sensors).
		In this paper, we introduce a novel algorithm by \textit{Cascading Convolutional Color Constancy} (in short, C$^4$) to improve robustness of regression learning and achieve stable generalization capability across datasets (different cameras and scenes) in a unique framework. The proposed C$^4$ method ensembles a series of dependent illumination hypotheses from each cascade stage via introducing a weighted multiply-accumulate loss function, which can inherently capture different modes of illuminations and explicitly enforce coarse-to-fine network optimization. Experimental results on the public Color Checker and NUS 8-Camera benchmarks demonstrate superior performance of the proposed algorithm in comparison with the state-of-the-art methods, especially for more difficult scenes.
	\end{abstract}
	
	\section{Introduction}\label{sec:introduction}
	The colors present in images are biased by the illumination in addition to the intrinsic reflection properties of scene objects and extrinsic spectral sensitivity across cameras, but they appear to be relatively constant for human visual perception systems.
	Such a property, referred to as color constancy, makes object appearance under diverse lighting sources independent of the casting illumination, which is desired in a large number of high-level vision problems.
	The color constancy problem can typically be addressed via estimating the color of the illuminant of the scene firstly, which then recovers the canonical colors of scene objects.
	A large number of computational color constancy algorithms \cite{qian2019finding,chen2019cumulative,cheng2015effective,bianco2017single,shi2016deep,hu2017fc4} rely on accurate and robust illumination predictions and then employ the simple yet effective von Kries model \cite{kries1902} for image correction.
	Estimating the illumination of an image can be formulated into learning a regression mapping from the imagery representation to its corresponding illumination label.
	Searching and identifying the best hypothesis of the illumination is not trivial in view of appearance inconsistency and label ambiguity.
	\begin{figure}[t]
		\centering \includegraphics[width=0.95\columnwidth]{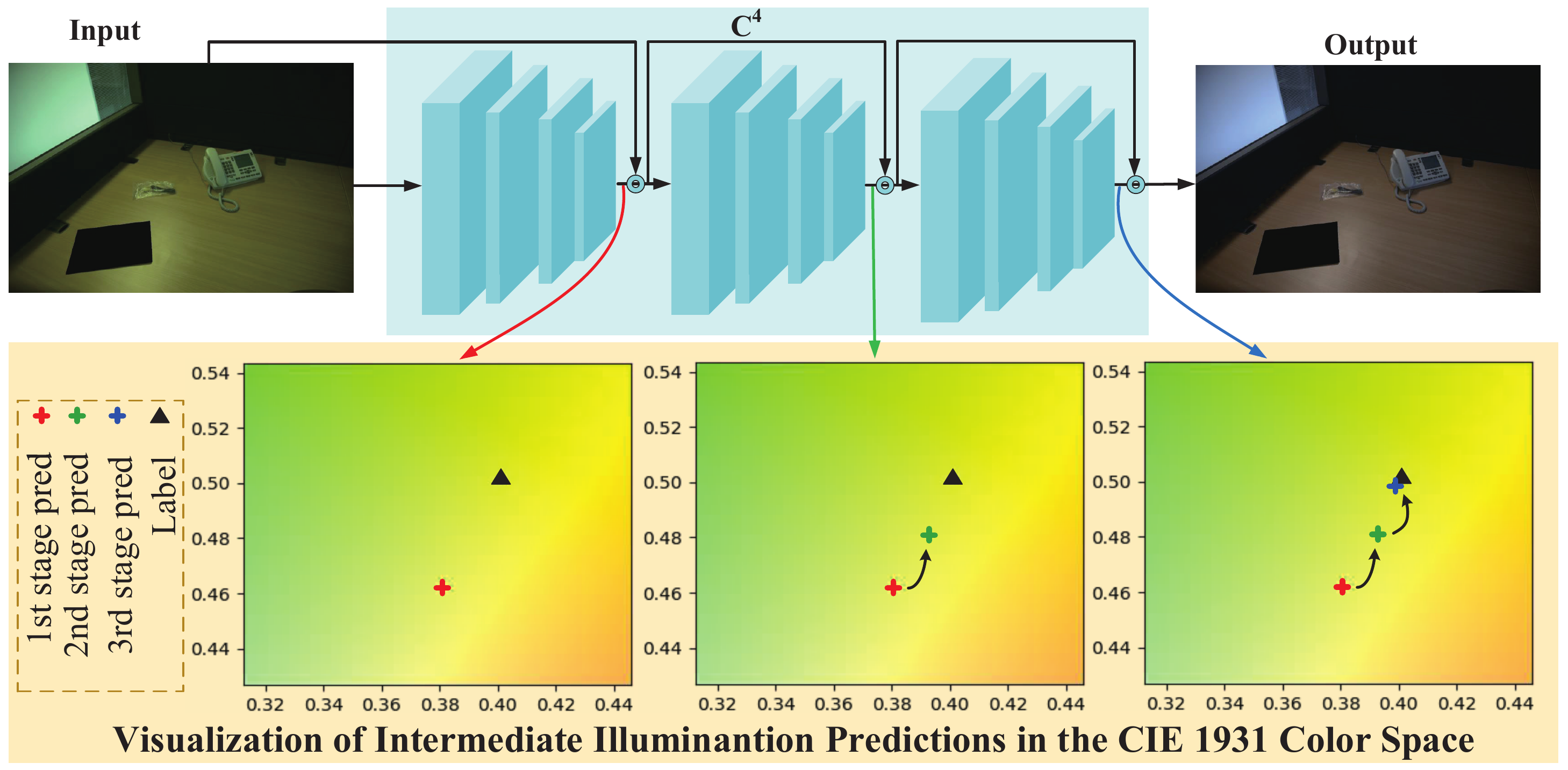}
		\caption{Visualization of the proposed C$^{4}$ method in a cascaded structure in the top row, while plots in the bottom row show dependent illumination hypotheses of different cascade stages in our C$^{4}$ on an example from the Color Checker dataset.
			Our C$^{4}$ can significantly boost illumination estimation performance in a coarse-to-fine refining manner. More examples are given in the experiment section.
		}
		\label{fig:intro}
	\end{figure}
	In addition to unknown surface reflection, the large appearance variation of the captured scene objects can be caused by the sensor sensitivity and also the illuminant spectrum.
	Specifically, spectral responses of sensors in cameras for color imaging are not consistent across camera models and brands, \textit{e}.\textit{g}. in the NUS 8-camera dataset \cite{cheng2014illuminant} one scene is captured with eight different cameras, and they have visually varying colors for the identical object surface. 
	Consequently, a typical solution is to train camera-specific estimators, which is less efficient and even impractical due to data-demanding characteristics.
	Very few algorithms \cite{qian2019finding} focus on the challenging camera-agnostic illumination estimation, achieving robust performance. Therefore, the challenge still exits.
	Most of the existing algorithms \cite{bianco2017single,barron2015convolutional,shi2016deep,barron2017fast,hu2017fc4} have been proposed to deal with appearance inconsistency, while very few concern on the challenge caused by the error-prone assumption in practice, \textit{i}.\textit{e}. one unique spectral illumination exists in the whole scene of each image. 
	In the procedure of label acquisition for color constancy datasets, a Macbeth ColorChecker chart is usually placed in the image, whose colors are recorded as the ground truth illumination, breaking the guarantee: the recorded ``ground truth'' represents the real global illumination.
	As a result, the gap between the label and the true scene illumination over spatial regions makes learning a regression more challenging, especially considering data augmentation via patch-based sampling widely adopted in state-of-the-art deep methods \cite{bianco2017single,shi2016deep,hu2017fc4}.
	Robustness against object appearance inconsistency and label ambiguity are desired imagery representation properties to learn from imagery observations and illumination labels.
	To achieve these, we introduce a multiply-accumulate loss function for cascading convolutional color constancy (\textit{e}.\textit{g}. FC$^4$ \cite{hu2017fc4} in the experiments) to cope with both challenges simultaneously. 
	In details, a series of dependent illumination hypotheses, reflecting different modes of illuminations, are generated via the proposed cascaded model, which are then combined in an ensemble to enforce explicitly coarse-to-fine refinement on illumination hypotheses as Figure \ref{fig:intro} shows. 
	The contributions of this paper are three-fold.
	\begin{itemize}
		\item This paper proposes a generic cascaded structure (\textit{i}.\textit{e}. the multiply-accumulate cascade) on illumination estimation to 1) ensemble multiple dependent illumination hypotheses and 2) achieve coarse-to-fine refinement, via a novel multiply-accumulate loss, which can be readily plugged into other learning-based illumination estimation methods.
		\item The proposed C$^{4}$ method increases model flexibility via enriching abstract features in a deeper network structure and also discovers latent correlation in the hypothesis space, which alleviates the suffering from ambiguous training samples.
		\item Extensive experiments on two popular benchmarks show that our C$^{4}$ achieves significantly better performance than the state-of-the-art, especially when coping with more difficult scenes.
	\end{itemize}
	Source codes and pre-trained models are available at {https://github.com/yhlscut/C4}.
	
	\section{Related work}
	\begin{figure*}[t]
		\centering \includegraphics[width=0.95\textwidth]{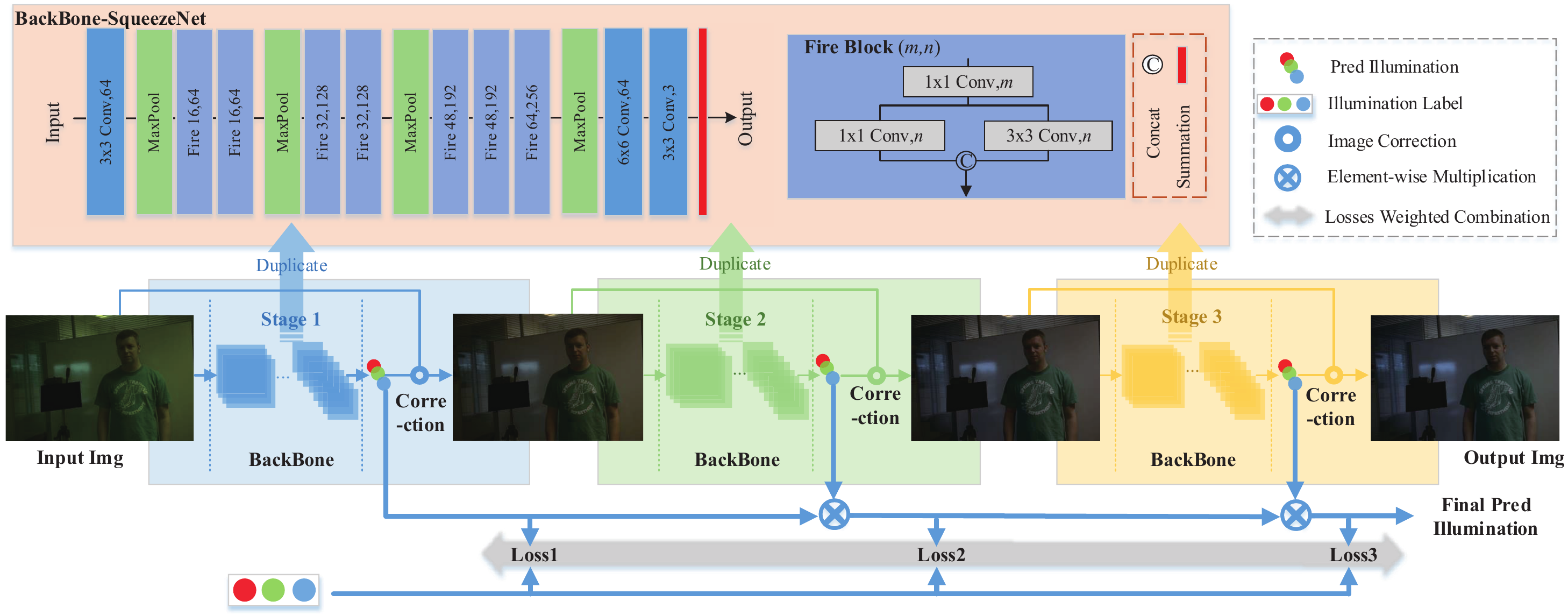}
		\caption{Pipeline of our three-stage C$^{4}$ model based on the SqueezeNet backbone. }\label{fig:pipeline}
	\end{figure*}
	Color constancy has been investigated for decades and numerous conventional algorithms are based on low-level imagery statistics, such as White-Patch \cite{brainard1986analysis}, Gray-World \cite{buchsbaum1980spatial}, Gray-Edge \cite{van2007edge}, Shades-of-Gray \cite{finlayson2004shades}, Bright Pixels \cite{joze2012role}, Grey Pixel \cite{yang2015efficient} and Gray Index \cite{qian2019finding}.
	These algorithms are proposed to determine the neutral white color with algorithm-specific assumptions, which encourage direct application to testing images in a learning-free fashion but can be sensitive in practice in consideration of their dependency on statistical distribution of pixel-wise colors, \textit{e}.\textit{g}. lack of gray pixels with using grey pixels \cite{yang2015efficient} and state-of-the-art statistical grey index \cite{qian2019finding}.
	
	Learning-based methods are a powerful alternative for generating constant colors under a scene  illumination, which can be categorized into two groups -- gamut mapping \cite{barnard2000improvements,chakrabarti2011color} and regression learning \cite{funt2004estimating,cheng2015effective,qian2017recurrent,chen2019cumulative,cardei1999committee,schaefer2005combined,bianco2017single,barron2015convolutional,shi2016deep,barron2017fast,hu2017fc4}.
	The former gamut mapping algorithms including edge-based \cite{barnard2000improvements}, intersection-based \cite{chakrabarti2011color} and pixels-based \cite{chakrabarti2011color} assume the size of colors under a given illuminant is limited, but will have a variation on observed colors when a deviation in the color of illuminants. 
	Given sufficient labeled training data, a model can be trained to recognize the canonical illumination by mapping from a gamut of a testing image under an unknown illuminant to the canonical gamut, which can thus generate an estimation of the scene illumination. 
	
	The latter regression learning-based algorithms aim to learn
	a direct regression mapping from the imagery representation to its corresponding illumination vector.  
	These methods focus on either designing robust regressors against large feature variation, based on support vector regression \cite{funt2004estimating}, regression trees \cite{cheng2015effective}, an ensemble of shallow regressors \cite{cardei1999committee,schaefer2005combined} or mining inter-dimensional label correlation as structured-output regression \cite{qian2016deep,chen2019cumulative}. 
	Inspired by the recent success of convolutional neural networks on numerous vision tasks,
	a number of works introduce the 2D convolutional feature encoding into color constancy. 
	\cite{bianco2017single} is the first attempt of deep color constancy, which copes with data sparsity problem demanded by fitting millions of network parameters via patch-based sampling.
	Convolutional Color Constancy (CCC) \cite{barron2015convolutional} and Fast Fourier Color Constancy (FFCC) \cite{barron2017fast} formulate the problem into a 2D spatial localization task on a 2D log-chroma space, while the difference of both methods lies in better performance and acceleration of the latter benefiting from extra semantic features and the BVM estimation in the frequency domain.
	In \cite{hu2017fc4}, a confidence-pooling layer is introduced to automatically feature encoding and discover the location of essential spatial regions for illumination estimation.   
	Existing deep learning methods mainly focus on designing network structure for robust feature encoding against the challenge of inconsistent appearance, but omits to benefit from combining multiple illumination hypotheses in an ensemble to handle ambiguous samples.
	
	Recent DS-Net \cite{shi2016deep} has two expert branches for first generating two hypotheses and then automatically selecting the better one.
	Similar to our C$^{4}$ method, it's motivation is to exploit multiple hypotheses of scene illumination for robust color constancy.
	However, there are two key differences.
	Firstly, the DS-Net conducts a discriminative selection instead of jointly learning to discover latent dependency across multiple illumination hypotheses as our C$^{4}$ model.
	Secondly, the DS-Net generates multiple independent illumination hypotheses in parallel, while the proposed C$^{4}$ method in a cascading network structure generates dependent hypotheses in serial to explicitly enforce coarse-to-fine refinement. 
	Experiment results in Tables \ref{tab:comparative} and \ref{tab:camera} demonstrate the superiority of our C$^{4}$ model to the DS-Net and other state-of-the-art methods.
	
	\section{C$^4$: Cascading Convolutional Color Constancy}
	\begin{table*}
		\begin{center}
			\caption{Comparative evaluation on two popular benchmarks. All results reported in this table are in units of degrees.}\label{tab:comparative}
			\resizebox{.95\textwidth}{!}{
				\begin{tabular}{l|rrrrr|rrrrr}
					\hline
					\multirow{2}{*}{Methods} & \multicolumn{5}{c|}{NUS 8-Camera \cite{cheng2014illuminant}} & \multicolumn{5}{c}{Color Checker \cite{shi2000re}}\\
					\cline{2-11}
					& Mean & Median & Tri-mean & Best 25\% & Worst 25\% & Mean & Median & Tri-mean &  Best 25\% & Worst 25\%  \\
					\hline
					\multicolumn{11}{l}{Static Methods} \\
					\hline
					White-Patch \cite{brainard1986analysis} & 10.62 & 10.58 & 10.49 & 1.86 & 19.45  & 7.55 & 5.68 & 6.35 & 1.45 & 16.12\\
					Gray-World \cite{buchsbaum1980spatial} & 4.14 & 3.20 & 3.39 & 0.90 & 9.00& 6.36 & 6.28 & 6.28 & 2.33 & 10.58  \\
					1st-order Gray-Edge \cite{van2007edge} & 3.20 & 2.22 & 2.43 & 0.72 & 7.69& 5.33 & 4.52 & 4.73 & 1.86 & 10.03\\
					2nd-order Gray-Edge \cite{van2007edge} & 3.20 & 2.26 & 2.44 & 0.75 & 7.27& 5.13 & 4.44 & 4.62 & 2.11 & 9.26 \\
					Shades-ofq-Gray \cite{finlayson2004shades}& 3.40 & 2.57 & 2.73 & 0.77 & 7.41& 4.93 & 4.01 & 4.23 & 1.14 & 10.20 \\
					General-Gray-World \cite{barnard2002comparison}& 3.21 & 2.38 & 2.53 & 0.71 & 7.10& 4.66 & 3.48 & 3.81 & 1.00 & 10.09\\
					Bright Pixels \cite{joze2012role}& 3.17 & 2.41 & 2.55 & 0.69 & 7.02  & 3.98 &  2.61 & - & - & - \\
					Cheng et al.2104 \cite{cheng2014illuminant}& 2.92 & 2.04 & 2.24 & 0.62 & 6.61 & 3.52 & 2.14 & 2.47 & 0.50 & 8.74\\
					LSRS \cite{gao2014efficient} & 3.45 & 2.51 & 2.70 & 0.98 & 7.32  & 3.31 & 2.80 & 2.87 & 1.14 & 6.39 \\
					Grey Pixel (edge) \cite{yang2015efficient}& 3.15 & 2.20 & - & - & -& 4.60 & 3.10 & - & - & - \\
					GI \cite{qian2019finding}& 2.91 & 1.97 & 2.13 & 0.56 & 6.67 & 3.07 & 1.87 & 2.16 & 0.43 & 7.62 \\
					\hline
					\multicolumn{6}{l}{Learning-based Methods} \\
					\hline
					Edge-based Gamut \cite{barnard2000improvements} & 8.43 & 7.05 & 7.37 & 2.41 & 16.08& 6.25 & 5.04 & 5.43 & 1.90 & 13.58 \\
					Bayesian \cite{gehler2008bayesian}& 3.67 & 2.73 & 2.91 & 0.82 & 8.21 & 4.82 & 3.46 & 3.88 & 1.26 & 10.49  \\
					MvCA \cite{chen2019cumulative} & - & - & - & - & - & 4.10 & 2.60 & - & - & - \\
					Intersection-based Gamut \cite{chakrabarti2011color}& 7.20 & 5.96 & 6.28 & 2.20 & 13.61& 4.20 & 2.39 & 2.93 & 0.51 & 10.70 \\  
					Pixels-based Gamut \cite{chakrabarti2011color}& 7.70 & 6.71 & 6.90 & 2.51 & 14.05& 4.20 & 2.33 & 2.91 & 0.50 & 10.72 \\
					Natural Images Statistics \cite{gijsenij2010color}& 3.71 & 2.60 & 2.84 & 0.79 & 8.47& 4.19 & 3.13 & 3.45 & 1.00 & 9.22 \\
					Spatio-spectral (GenPrior) \cite{chakrabarti2011color}& 2.96 & 2.33 & 2.47 & 0.80 & 6.18& 3.59 & 2.96 & 3.10 & 0.95 & 7.61 \\
					Corrected-Moment$^1$ (19 Color) \cite{finlayson2013corrected} & 3.05 & 1.90 & 2.13 & 0.65 & 7.41& 2.96 & 2.15 & 2.37 & 0.64 & 6.69 \\
					Corrected-Moment$^1$ (19 Edge) \cite{finlayson2013corrected}& 3.03 & 2.11 & 2.25 & 0.68 & 7.08& 3.12 & 2.38 & 2.59 & 0.90 & 6.46 \\
					Exemplar-based \cite{joze2013exemplar} & - & - & - & - & -& 3.10 & 2.30 & - & - & -\\
					Chakrabarti et al. 2015 \cite{chakrabarti2015color} & - & - & - & - & -  & 2.56 & 1.67 & 1.89 & 0.52 & 6.07\\
					Regression Tree \cite{cheng2015effective}& 2.36 & 1.59 & 1.74 & 0.49 & 5.54& 2.42 & 1.65 & 1.75 & 0.38 & 5.87 \\
					CNN \cite{bianco2017single}& - & - & - & - &-& 2.36 & 1.98 & - & - & - \\
					CCC (dist+ext) \cite{barron2015convolutional} & 2.38 & 1.48 & 1.69 & 0.45 & 5.85& 1.95 & 1.22 & 1.38 & 0.35 & 4.76  \\
					DS-Net (HypNet+SeNet) \cite{shi2016deep}& 2.24 & 1.46 & 1.68 & 0.48 & 6.08& 1.90 & 1.12 & 1.33 & 0.31 & 4.84  \\
					FFCC \cite{barron2017fast}& 1.99 & \bf{1.31} & \bf{1.43} & \bf{0.35} & 4.75  & 1.78& 0.96 &1.14 &0.29 &4.62 \\
					AlexNet-FC$^4$ \cite{hu2017fc4}& 2.12 & 1.53 & 1.67 & 0.48 & 4.78 & 1.77 & 1.11 & 1.29 & 0.34 & 4.29 \\
					SqueezeNet-FC$^4$ \cite{hu2017fc4} & 2.23 & 1.57 & 1.72 & 0.47 & 5.15& 1.65 & 1.18 & 1.27 & 0.38 & 3.78\\
					\hline
					C$^{4}$$_\text{AlexNet-FC4}$ (ours)  & 2.07 & 1.47 & 1.63 & 0.48 & 4.63& 1.49 & 1.03 & 1.13 & 0.29 & 3.52\\
					C$^{4}$$_\text{SqueezeNet-FC4}$ (ours) & \bf{1.96} & 1.42 & 1.53 & 0.48 & \bf{4.40}& \bf{1.35} & \bf{0.88} & \bf{0.99} & \bf{0.28} & \bf{3.21} \\
					\hline
				\end{tabular}
			}
		\end{center}
	\end{table*} 
	The problem definition of single illumination estimation problem is to predict the illumination vector $\bm{y}\in\mathbb{R}^3$ from the image $X\in\mathbb{R}^{H\times W\times 3}$.
	For learning-based illumination estimation, 
	the objective function can be written as the following:
	\begin{equation}
	\min_{\theta}~~ \mathcal{L}({f}^{\theta}(X), \bm{y}), \label{eqn:obj}
	\end{equation}
	where ${f}^{\theta}(\cdot) \in \mathbb{R}^3$ is the mapping from the image  ${X}$ to illumination vector $\bm{y}$, and $\theta$ denotes the model parameters of $f$ to be optimized.
	$\mathcal{L}(\cdot)$ denotes the loss function and the typical loss in illumination estimation is the angular loss (formulated in Equation~(\ref{eqn:loss})).
	During testing, given an input, the trained model $f^{\theta}(\cdot)$ infers the predicted illumination $f^{\theta}(X)$, which is used to generate the color-corrected image.
	In the context of convolutional color constancy, $f^{\theta}(\cdot)$ is the output of a deep network, while $\theta$ denotes the network weights.
	This section will present an overview of the proposed C$^4$ algorithm, a novel multiply-accumulate loss, image correction, and implementation details respectively.   
	
	\subsection{Network Structure}\label{subsec.cov_ill_est}
	The C$^{4}$ network consists of three stages as illustrated in Figure \ref{fig:pipeline}.
	Given training pairs $\{X, \bm{y}\}_i$, $i\in \{1,2,\cdots,N$\}, 
	in a cascaded structure, ${f}^{\theta}(\cdot)$ can be decomposed into ${f}_l(\cdot), l =1,2,\ldots, L$, where $l$ and $L$ denote the cascade level and the total number of cascaded stages, respectively, with $\theta$ omitted for simplicity. 
	We define $f_{l}f_{l-1}(X)$ as a simpler notation for $f_{l}(X / f_{l-1}(X) )$ (image correction, explained in Equation (\ref{eq:correction})). 
	Considering the cascaded structure, now Equation (\ref{eqn:obj}) of the three-stage C$^{4}$ can be written as the following:
	\begin{equation}
	\min_{\theta}~~ \mathcal{L}(f_3f_2f_1(X)), \bm{y}; \theta). \label{eqn:obj_cascade}
	\end{equation} 
	
	In the light of its good performance in illumination estimation, we employ the state-of-the-art CNN model -- FC$^4$ based on the AlexNet and SqueezeNet backbone in \cite{hu2017fc4}. 
	In details, the FC$^4$ adopts low-level convolutional layers of off-the-shelf AlexNet and SqueezeNet pre-trained on the ImageNet \cite{deng2009imagenet}, and replaces the remaining layers with two more convolutional layers.
	Specifically, the AlexNet-FC$^4$ model keeps all the layers up to the \textit{conv5} layer and replaces the rest fully-connected layers with \textit{conv6} having $6\times 6\times 64$ convolutional filters and \textit{conv7} ($1\times 1\times 4$), while the detailed network structure of the SqueezeNet-FC$^4$ is shown in Figure \ref{fig:pipeline}.
	For both networks, every convolution layers are followed by a ReLU non-linearity, and a dropout with probability $0.5$ is added before the last convolutional layer. 
	It is noted that a confidence-weighted pooling layer is followed by the last \textit{conv} layer in original FC$^4$ to improve robustness against color consistency across spatial regions via suppressing less confident predictions, while our FC$^4$ model employs a much simpler summation on the output of last \textit{conv} layer to obtain global illumination $\bm{y}$ (\textit{i}.\textit{e}. a red bar in the top row of Figure \ref{fig:pipeline}) without hindering the performance.
	
	\subsection{A Novel Multiply-Accumulate Loss}\label{subsec.loss}
	As mentioned earlier, illumination predictions in different cascade stages are all approximation to ground truth illumination, which can be viewed as its different nodes.
	Different from the DS-Net \cite{shi2016deep} to design a selection mechanism via training another branch to determine the better hypothesis, the proposed cascaded network aims to exploit latent dependency across illumination hypotheses to explicitly enforce coarse-to-fine refinement approaching the ground truth.
	To this end, we introduce a combined multiply-accumulate loss on all hypotheses to capture their latent correlation to refine illumination hypotheses,which is depicted as the following equation:
	\begin{equation}
	\mathcal{L} =
	\sum_{l=1}^{L}  
	\mathcal{L}^{(l)}(\prod_{i=1}^{l}{f}_i(X_i),\bm{y}) \\
	\label{eqn:loss}
	\end{equation}
	where $\mathcal{L}^{(l)}$ represents the loss at the $l$-th cascade stage. 
	Moreover, the proposed loss can alleviate cumulative errors via supervision on intermediate illumination predictions.  
	We also consider its simple weighted extension as
	\begin{equation}
	\mathcal{L} =
	\sum_{l=1}^{L}  
	w_l\mathcal{L}^{(l)}(\prod_{i=1}^{l}{f}_i(X_i),\bm{y}) \\
	\label{eqn:wloss}
	\end{equation}
	where $w_l$ denotes weights for the loss on illumination prediction in the $l$-th stage and ground truth $\bm{y}$. We compare the variants of weights in Equation (\ref{eqn:wloss}) and results are shown in Table \ref{tab:weight}.
	The proposed losses are embedded into the deep cascaded network in an end-to-end learning manner as shown in Figure \ref{fig:pipeline}.
	
	For large appearance variation and ambiguous labels, a selection or an ensemble of a number of illumination estimators are verified its superior robustness, but it remains challenging to capture latent correlation across illumination hypotheses. The combined loss proposed in this paper is extremely simple yet effective, as the principle of our design can be explained by enforcing each cascaded stage to learn a specific correction pattern to suppress ambiguous hypotheses in previous stages.      
	\begin{table*}
		\begin{center}
			\caption{Camera-agnostic evaluation. All results are in units of degrees.}\label{tab:camera}
			\resizebox{.95\textwidth}{!}{
				\begin{tabular}{l|rrrrr|rrrrr}
					\hline
					Training set & \multicolumn{5}{c|}{NUS 8-Camera} & \multicolumn{5}{c}{Color Checker}\\
					Testing set & \multicolumn{5}{c|}{Color Checker} & \multicolumn{5}{c}{NUS 8-Camera} \\
					& Mean & Median & Tri-mean &  Best 25\% & Worst 25\% & Mean & Median & Tri-mean & Best 25\% & Worst 25\%  \\
					\hline
					\multicolumn{11}{l}{Static Methods} \\
					\hline
					White-Path \cite{brainard1986analysis} & 7.55 & 5.68 & 6.35 & 1.45 & 16.12 & 9.91 & 7.44 & 8.78 & 1.44 & 21.27 \\
					Gray-World \cite{buchsbaum1980spatial} & 6.36 & 6.28 & 6.28 & 2.33 & 10.58 & 4.59 & 3.46 & 3.81 & 1.16 & 9.85 \\
					1st-order Gray-Edge \cite{van2007edge} & 5.33 & 4.52 & 4.73 & 1.86 & 10.43 & 3.35 & 2.58 & 2.76 & 0.79 & 7.18 \\
					2nd-order Gray-Edge \cite{van2007edge} & 5.13 & 4.44 & 4.62 & 2.11 & 9.26 & 3.36 & 2.70 & 2.80 & 0.89 & 7.14 \\
					Shades-of-Gray \cite{finlayson2004shades} & 4.93 & 4.01 & 4.23 & 1.14 & 10.20 & 3.67 & 2.94 & 3.03 & 0.99 & 7.75 \\
					General-Gray-World \cite{barnard2002comparison} & 4.66 & 3.48 & 3.81 & 1.00 & 10.09 & 3.20 & 2.56 & 2.68 & 0.85 & 6.68 \\
					Grey Pixel (edge) \cite{yang2015efficient} & 4.60 & 3.10 & - & - & - & 3.15 & 2.20 & - & - & - \\
					Cheng et al. 2104 \cite{cheng2014illuminant} & 3.52 & 2.14 & 2.47 & 0.50 & 8.74 & 2.92 & 2.04 & 2.24 & 0.62 & 6.61 \\
					LSRS \cite{gao2014efficient} & 3.31 & 2.80 & 2.87 & 1.14 & 6.39 & 3.45 & 2.51 & 2.70 & 0.98 & 7.32 \\
					GI \cite{qian2019finding} & 3.07 & \bf{1.87} & \bf{2.16} & \bf{0.43} & 7.62 & 2.91 & 1.97 & 2.13 & \bf{0.56} & 6.67 \\
					\hline
					\multicolumn{11}{l}{Learning-based Methods} \\
					\hline
					Bayesian \cite{gehler2008bayesian} & 4.75 & 3.11 & 3.50 & 1.04 & 11.28 & 3.65 & 3.08 & 3.16 & 1.03 & 7.33 \\
					Chakrabarti et al. 2015 \cite{chakrabarti2015color} & 3.52 & 2.71 & 2.80 & 0.86 & 7.72 & 3.89 & 3.10 & 3.26 & 1.17 & 7.95 \\
					FFCC \cite{barron2017fast} & 3.91 & 3.15 & 3.34 & 1.22 & 7.94 & 3.19 & 2.33 & 2.52 & 0.84 & 7.01 \\
					\text{AlexNet}-FC$^4$  \cite{hu2017fc4} & 3.23 & 2.57 & 2.73 & 0.90 & 6.70 & 2.62 & 2.16 & 2.25 & 0.79 & 5.23 \\
					\text{SqueezeNet}-FC$^4$ \cite{hu2017fc4} & 3.02 & 2.36 & 2.50 & 0.81 & 6.36 & 2.40 & 2.03 & 2.10 & 0.70 & 4.80\\
					C$^{4}_\text{AlexNet-FC4}$ (ours) & 2.85 & 2.26 & 2.38 & 0.76 & 5.97 & 2.52 & 2.07 & 2.15 & 0.69 & 5.20 \\
					C$^{4}_\text{SqueezeNet-FC4}$ (ours) & \bf{2.73} & 2.20 & 2.28 & 0.72 & \bf{5.69} & \bf{2.28} & \bf{1.90} & \bf{1.97} & 0.67 & \bf{4.60}\\
					\hline
			\end{tabular}}
		\end{center}
	\end{table*}
	\begin{figure}[thbb]
		\centering \includegraphics[width=0.95\columnwidth]{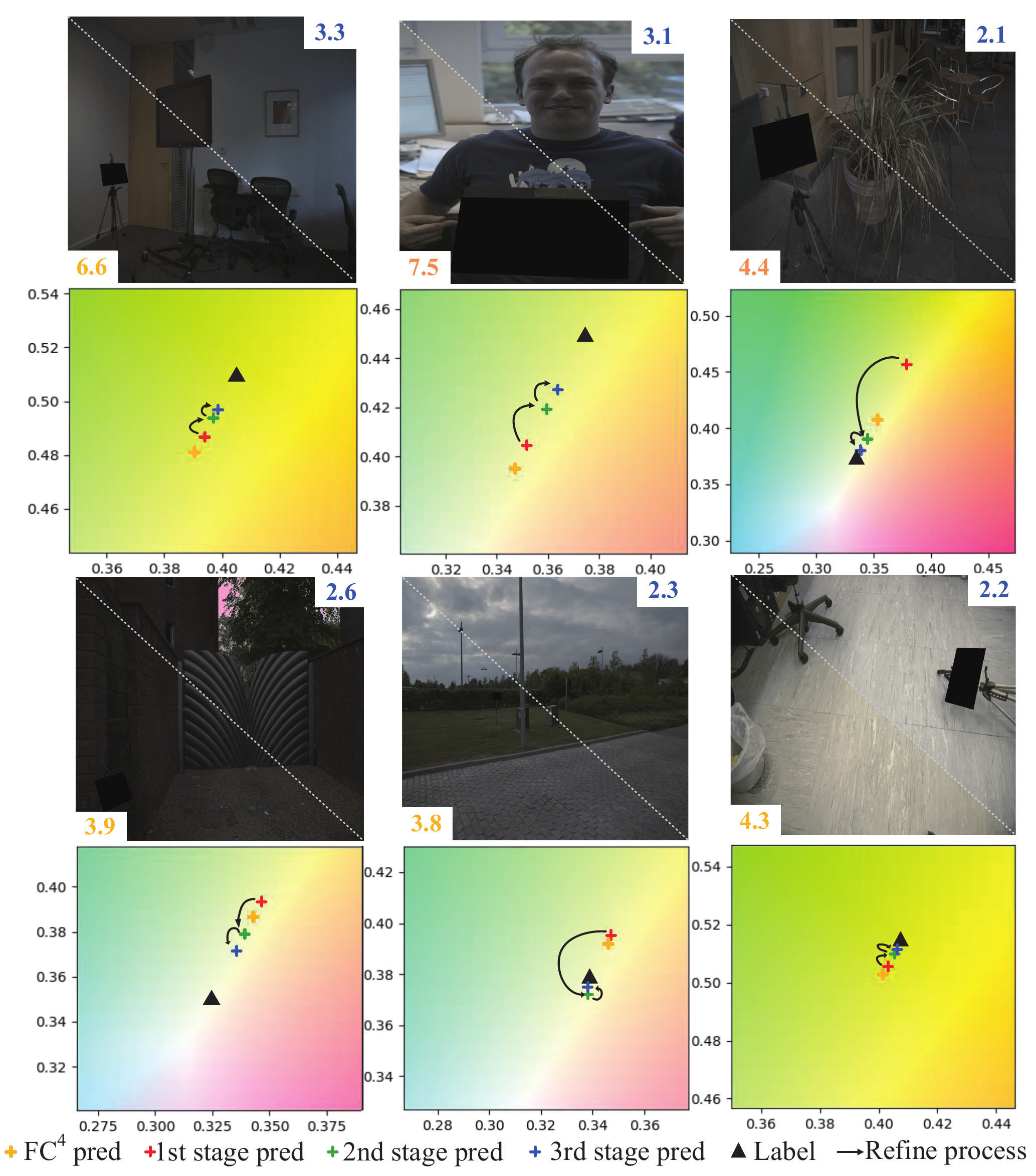}
		\caption{Visualization of harder samples from the Color Checker dataset. In the 1st and 3rd rows, the lower left parts of images are corrected by our detected hypotheses and the other parts are images corrected by the FC$^{4}$'s predictions. The numbers in the white rectangles of pictures are angle errors (in degrees) between illumination predictions and labels. The 2nd and 4th rows show the trajectories of predictions towards ground truth labels. }
		\label{fig:harder}
	\end{figure}
	
	\subsection{Image Correction}
	With an estimated illumination $\hat{\bm{y}} = [\hat{y}_{r},\hat{y}_{g},\hat{y}_{b}] \in \mathbb{R}^3$ for a biased image $\bm{X}$ with the trained C$^4$ model, the canonical colors of scene objects in the image can be recovered under the simplified assumption that each RGB channel can be modified separately \cite{kries1902}.
	In other words, we can obtain the corrected image $\bar{\bm{X}}\in\mathbb{R}^{H\times W\times 3}$ under the canonical illumination as
	\begin{equation}
	\label{eq:correction}
	\bar{\bm{X}_{j}} = \bm{X}_{j}/{y_{j}} \in \mathbb{R}^{H\times W}, ~j\in \{R, G, B\}.
	\end{equation}
	
	\subsection{Implementation Details}\label{subsec.imple}
	In data augmentation, we randomly crop patches from original images with a side length of [$0.1$, $1$] times the shorter side of the original image which are randomly rotated between $-30^\circ$ and $30^\circ$. 
	These patches are then resized into 512 $\times$ 512 pixels and finally randomly horizontal flipped with a probability of $0.5$. 
	To increase the diversity of limited training data, the illumination labels in each image are scaled by three different random values within the range between 0.6 and 1.4, and pixel-wise scene colors present in the original image are also biased by the randomly generated ratios.  
	We further apply gamma correction to convert linear images into nonlinear images and normalize the values of the images to $[0,1]$.
	During training, the ADAM algorithm \cite{kingma2014adam} is employed to train the model with a fixed batch size (\textit{i}.\textit{e}. 16 in our experiments), and the learning rate is set to $3\times 10^{-4}$  and $1\times10^{-4}$ for our C$^{4}$ model based on the SqueezeNet and AlexNet backbone respectively. 
	For computational efficiency and robust performance, we first train the one-stage C$^{4}$ for $2k$ epochs, the learned network weights are loaded to each cascade stage as initial weights in our three-stage C$^{4}$ model for further fine-tuning jointly.
	\begin{figure*}[t]
		\centering \includegraphics[width=0.95\textwidth]{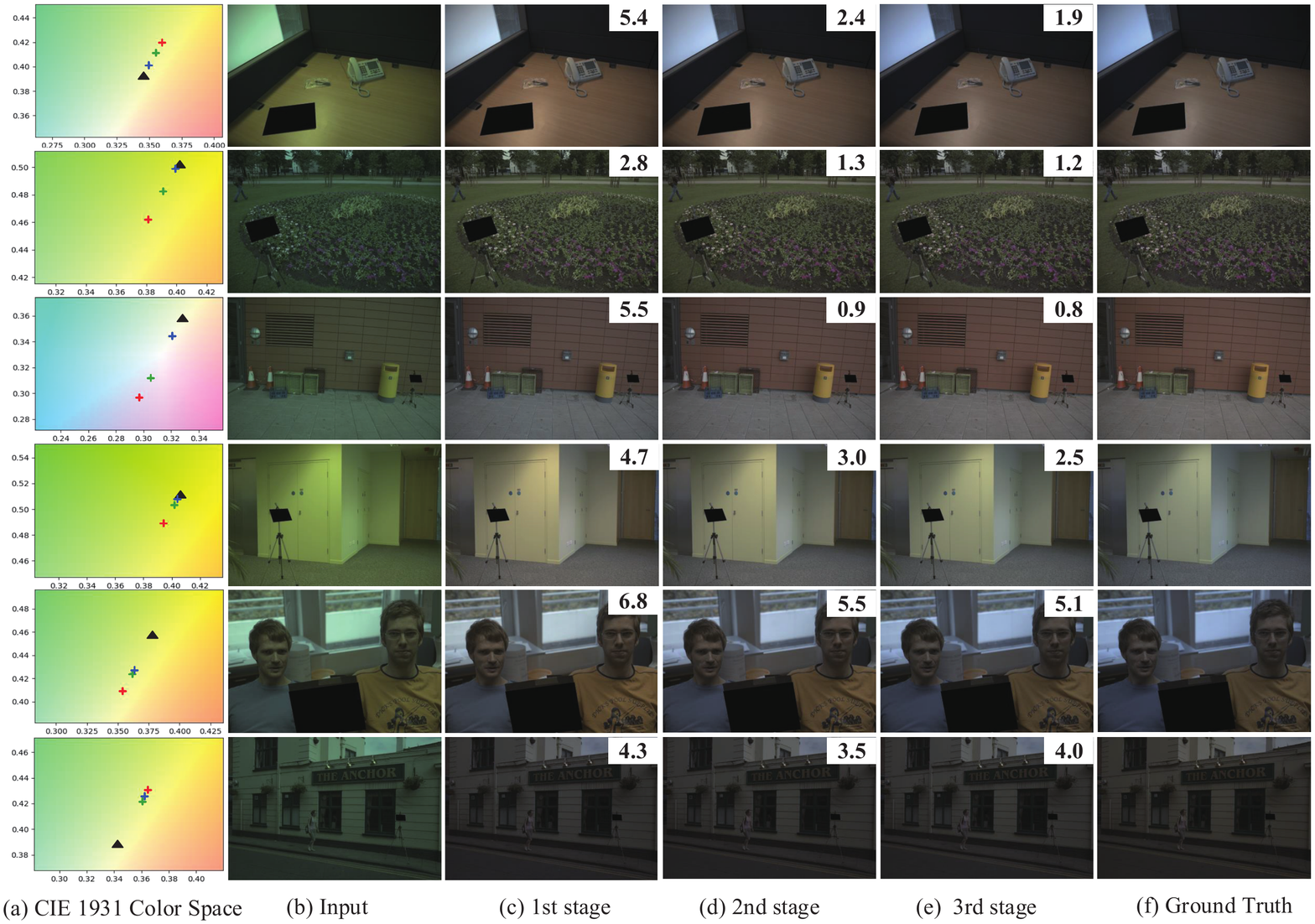}
		\caption{Visualization of a number of examples  from the Color Checker dataset. (a) In the CIE 1931 color space chromaticity diagram, where the red, green and blue plus sign '$+$' represent the 1st stage, the 2nd stage and the 3rd stage illumination predictions of our C$^{4}$ given input images in (b), the black triangle is the corresponding illumination labels. (c) (d) and (e) are corrected images by intermediate and final illumination hypotheses spotted by the C$^{4}$. The angle errors (in degrees) between illumination predictions and labels are highlighted in the white rectangles in the top-right of images.}
		\label{fig:comparison}
	\end{figure*}
	
	\section{Experiments}\label{sec.exper}
	\begin{figure}[t]
		\centering \includegraphics[width=0.95\columnwidth]{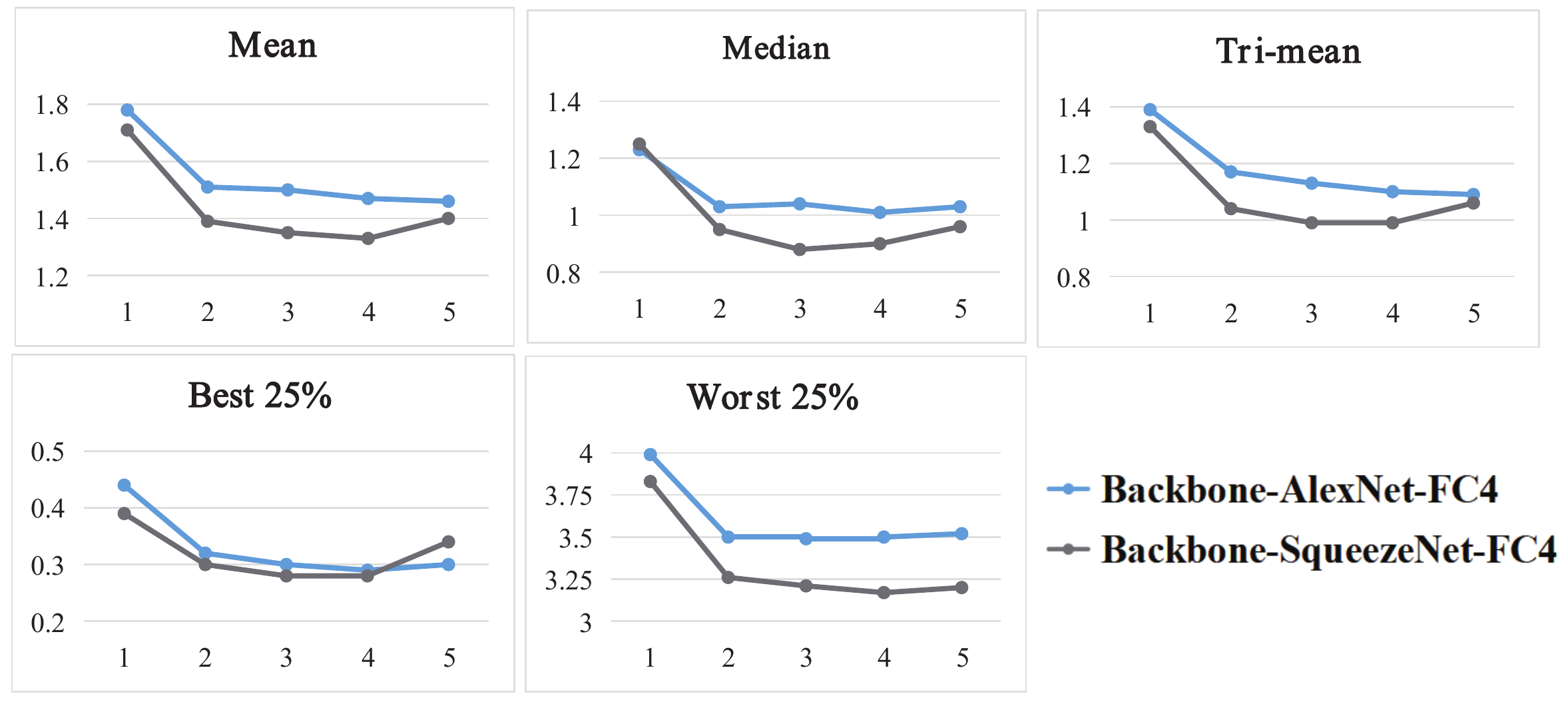}
		\caption{Evaluation on effect of cascade size on the Color Checker dataset. The x axis: the length of the cascade, while the y axis: the angular error (in degrees).}
		\label{fig:cascade_size}
	\end{figure}
	
	\subsection{Datasets and Settings}\label{subsec.dataset}
	We conduct experimental evaluation on two public color constancy benchmarks: the NUS 8-Camera dataset \cite{cheng2014illuminant} and the re-processed Color Checker dataset \cite{shi2000re}. 
	The NUS 8-camera dataset is composed of 1736 images from 8 commercial cameras, while the Color Checker dataset contains 568 images including indoor and outdoor scenes. 
	All images in both benchmarks are linear images in the RAW format of the acquisition device, each with a Macbeth ColorChecker (MCC) chart, which provides an estimation of illuminant colors.
	
	To prevent the convolutional network from detecting and utilizing MCCs as a visual cue, all images are masked with provided locations of MCC during training and testing. 
	Following \cite{chen2019cumulative,qian2019finding,barron2015convolutional}, we adopt three-fold cross-validation on both datasets in all experiments. 
	
	As suggested in \cite{hordley2004re} as well as a number of recent works \cite{chen2019cumulative,qian2019finding,barron2015convolutional}, we use the \textit{angular error} $\epsilon$ between the RGB triplet of estimated illuminant $\hat{\bm{y}}$ and the RGB triplet of the measured ground truth illuminant $\bm{y}$ as the performance metric denoted as :
	\begin{equation}
	\epsilon(\hat{\bm{y}},\bm{y})  = \arccos \left(\frac{\hat{\bm{y}} \cdot \bm{y}}{\|\hat{\bm{y}} \| \|\bm{y} \|}\right);
	\end{equation}
	where $\cdot$ denotes the inner product between vectors, $\| \cdot \|$ is the Euclidean norm. In our experiments, the mean, median, tri-mean of all the angular errors, mean of the best $25\%$ and the worst $25\%$ errors are reported. 
	
	\subsection{Comparison to State-of-the-Art Methods} \label{subsec.compar}
	Table \ref{tab:comparative} compares the proposed C$^{4}$ with the state-of-the-art methods in terms of the Mean, Median, Tri-mean, the Best $25\%$ and the Worst $25\%$ of angular errors on two datasets. 
	The proposed method can beat most of color constancy algorithms except the FFCC \cite{barron2017fast}.
	{{On the one hand, on the Color Checker dataset, our method significantly outperforms the FFCC on all five metrics, especially there are $18.18\%$ and $15.10\%$ improvements in the Mean and the  Worst $25\%$ metrics.
On the other hand, on the NUS 8-Camera benchmark, although FFCC outperforms on some metrics, our C$^{4}_\text{SqueezeNet-FC4}$ is better than FFCC on the mean and worst $25\%$ metrics. 
	}}
Performance gap on the NUS 8-Camera can be explained by the limited size of scenes (i.e., each scene generates 8 images with different cameras) leading to less positive effects of data augmentation in our method. 	
	More importantly, the C$^{4}$ can consistently beat its direct competitors -- its backbone \text{AlexNet}-FC$^4$ and \text{SqueezeNet}-FC$^4$ in all five metrics on both datasets.
	In view of the identical network structure for feature encoding, performance gains can only be explained by the design of the cascaded network structure.
	It is noted that the proposed C$^4$ method consistently outperforms its backbone FC$^4$ in the more challenging scenes, as illustrated in Figure \ref{fig:harder}.
	
	\subsection{Evaluation on Camera-Agnostic Color Constancy}
	To verify the robustness of our model against appearance inconsistency due to camera sensitivity, we take two disjoint datasets, one for training and the other for testing.
	Specifically, we conduct an evaluation on the Color Checker dataset with a model trained on the NUS 8-camera dataset and vice versa, whose results are reported in Table \ref{tab:camera}.
	Compared to the state-of-the-art statistical GI \cite{qian2019finding}, the C$^{4}$ achieves competitive performance and even performs best in the Worst 25\% metric consistently and significantly.
	Moreover, our C$^{4}$ with different backbone CNNs achieve the best performance again among learning-based illumination estimation in all performance metrics on both datasets, which verifies that our model can mitigate negative effects of imaging patterns across cameras owing to its strong generalization capability via progressive refinement and data argumentation. 
	
	\subsection{Discussion about Loss Combination}
	In our cascaded structure, the combination of loss functions is something worth discussing. We further discuss the design of our loss function with two strategies: a single multiplication loss and the weighted multiply-accumulate loss, with the three-stage C$^4$ model.
	\begin{itemize}
		\item {\bf Single multiplication Loss --} It only penalizes the final fine illumination prediction. (\textit{e}.\textit{g}. in Equation (\ref{eqn:wloss}), when $L=3$, weights should be $[w_1,w_2,w_3] = [0,0,1]$) 
		\item {\bf Weighted multiply-accumulate loss --} It combines the intermediate illumination prediction from each stage, and penalizes these illumination hypotheses jointly. (\textit{e}.\textit{g}. in Equation (\ref{eqn:wloss}), when $L=3$, weights satisfying $w_1 \times w_2\times w_3\not=0$) 
	\end{itemize}
	Table \ref{tab:weight} shows comparative results on combined strategies of the loss function.
	The latter, the weighted multiply-accumulate loss Equation(\ref{eqn:wloss}) is superior to its specific case -- single multiplication loss, which supports our motivation to design the multiply-accumulate loss to exploit multiple illumination hypotheses.
	Moreover, among the settings of weights, the equal weight can be slightly better than the remaining, although the improvement is very marginal.
	
	\begin{table}[t]
		\centering
		\caption{Statistics of angular errors (in degrees) obtained by different loss combinations of the three-stage C$^{4}$ model on the Color Checker dataset.}\label{tab:weight}
		\resizebox{.95\columnwidth}{!}
		{
			\begin{tabular}{ccc|ccccc}
				\hline
				\multirow{2}{*}{$w_1$} & \multirow{2}{*}{$w_2$} & \multirow{2}{*}{$w_3$} &  \multirow{2}{*}{Mean} & \multirow{2}{*}{Median} & \multirow{2}{*}{Tri-mean} & Best & Worst\\
				
				& & & & & & 25\% &25\%\\			
				\hline
				\multicolumn{6}{l}{Backbone-SqueezeNet-FC$^4$} \\
				\hline
				0.00 & 0.00 & 1.00 & 1.48 & 0.97 & 1.10 & 0.32 & 3.50 \\
				0.20 & 0.30 & 0.50 & 1.37 & 0.92 & 1.03 & 0.29 & 3.26  \\
				0.33 & 0.33 & 0.34 &\bf{1.35} & \bf{0.88} & \bf{0.99} & \bf{0.28} & \bf{3.21} \\
				0.50 & 0.30 & 0.20 & 1.38 & 0.90 & 1.00 & 0.32 & 3.23 \\ 
				0.70 & 0.20 & 0.10 & 1.37 & 0.89 & 1.00 & 0.31 & 3.25 \\
				
				\hline
				\multicolumn{6}{l}{Backbone-AlexNet-FC$^4$} \\
				\hline
				0.00 & 0.00 & 1.00 & 1.57 & 1.09 & 1.22 & 0.32 & 3.60 \\
				0.20 & 0.30 & 0.50 & 1.52 & 1.07 & 1.17 & 0.32 & \bf{3.48} \\
				0.33 & 0.33 & 0.34 & \bf{1.49} & 1.03 & \bf{1.13} & \bf{0.29} & 3.52\\
				0.50 & 0.30 & 0.20 & 1.50 & \bf{1.01} & 1.14 & 0.33 & 3.50 \\
				0.70 & 0.20 & 0.10 & 1.50 & 1.02 & 1.14 & 0.32 & 3.50 \\
				\hline
			\end{tabular}
		}
	\end{table}
	\begin{table}[!t]
		\newcommand{\tabincell}[2]{\begin{tabular}{@{}#1@{}}#2\end{tabular}}
		\centering
		\caption{Comparison of network parameters on the Color Checker dataset. The model labeled ``1/3p'' indicates the backbone network parameters are reduced by one third. ``3-stage'' means three-stages model. C$^{4}_\text{B}$ and C$^{4}_\text{E}$ mean our proposed network with the model in B) and E) as the backbone respectively. All results in this table are in units of degrees.}\label{tab:parameters}
		\resizebox{.95\columnwidth}{!}
		{
			\begin{tabular}{l|ccccc}
				\hline
				\multirow{2}{*}{Method} & \multirow{2}{*}{Mean} & \multirow{2}{*}{Median} & \multirow{2}{*}{Tri-mean} & Best & Worst\\
				& & & & 25\% & 25\% \\
				\hline
				A) AlexNet-FC4 & 1.77 & \bf{1.11} & 1.29 & \bf{0.34} & 4.29 \\ 
				B) AlexNet-FC4,1/3p & 2.17 & 1.58 & 1.71 & 0.53 & 4.87\\ 
				C) C$^{4}_\text{B}$,3 stage & \bf{1.65} & \bf{1.11} & \bf{1.22} & \bf{0.34} & \bf{3.88}\\ 
				\hline
				D) SqueezeNet-FC4 & 1.65 & 1.18 & 1.27 & 0.38 & 3.78 \\
				E) SqueezeNet-FC4,1/3p & 1.94 & 1.40 & 1.52 & 0.49 & 4.31 \\
				F) C$^{4}_\text{E}$,3-stage & \bf{1.47} & \bf{0.97} & \bf{1.09} & \bf{0.31} & \bf{3.49} \\
				\hline
			\end{tabular}
		}
	\end{table}
	
	\subsection{Discussion of Cascade Size}
	Another key insight of our C$^{4}$ is to incrementally improve illumination predictions in a cascaded structure. 
	Performance of such a cascaded structure depends on the size of cascade stages. 
	We demonstrate the validity of our cascaded structure by comparing the performance at varying cascade levels.
	As shown in Figure \ref{fig:cascade_size}, angular errors in all metrics of two C$^{4}$ variants decrease with cascade level increasing. 
	In particular, the performance increases by a big margin from one-stage C$^{4}$ to two-stage variant, while a moderate improvement from two-stage to three-stage, or even to four-stage.
	However, as the number of cascades continues to increase, the performance does not improve. We suppose that a deeper network makes it harder to fit dramatically increasing size of network parameters.
	Such a phenomenon encourages a relatively large size of cascade stages for color constancy.
	
	To further illustrate the effectiveness of the introduced cascade structure, we visualize some examples with intermediate illumination predictions from each stage of the proposed C$^{4}$ cascade on the Color Checker dataset in Figure \ref{fig:comparison}.
	{{Most corrected images in (c) and (d) are visually closer to ground truth (GT) than those in (b), and we quantitatively measure predictions in the 1st, 2nd and 3rd stages of three-level C$^4$ model with ground truth of testing samples, $P(1,2)=69.72\%$ and $P(2,3)=60.21\%$, where $P(l,l+1)$ denotes the ratios of more accurate predictions of the $(l+1)$-th stage in comparison with those of the $l$-th stage during testing. 
	It further verifies the rationale of the coarse-to-fine cascade structure.}}
	
	\subsection{Evaluation with Comparable Network Parameters}
	As aforementioned, the performance of such a cascade structure can be improved with the increase of the size of the cascade stages $L$ (when $L<=4$). However, the number of network parameters is proportional to the size of $L$.
	To explore the real source of this improvement, we compress the number of network parameters of our backbone (\textit{i}.\textit{e}. AlexNet-FC4 and SqueezeNet-FC4) by decreasing the number of convolution kernels in every convolutional layer. 
	As shown in Table \ref{tab:parameters}, the numbers of network parameters in method B) and E) are one-third of original backbone method A) and D) after compressing. 
	When using compressed backbone networks, we get new cascade models (\textit{i}.\textit{e}. three-stage C$^4$ methods C) and F)), whose sizes of network parameters are comparable to original FC$^4$ models in A) and D).  
	Table \ref{tab:parameters} reveals that superior performance of our C$^4$ can be credited to the cascade network structure.
	
	\section{Conclusion}
	This paper designs a cascade of convolutional neural networks for color constancy, which consistently achieves the best performance for more challenging samples (on the Worst 25\% metric) and more robust performance under the camera-agnostic setting.
	Experiment results favor for a relatively larger cascade size and verify the boosting benefits of combining multiple illumination hypotheses and the coarse-to-fine refinement.
	
	\section{Acknowledgements}
	{{This work is supported in part by the National Natural Science Foundation of China (Grant No.: 61771201, 61902131), the Program for Guangdong Introducing Innovative and Enterpreneurial Teams (Grant No.: 2017ZT07X183), the Fundamental Research Funds for the Central Universities (Grant No.: D2193130), and the SCUT Program (Grant No.: D6192110).
	}}
	
	\bibliographystyle{aaai}
	\bibliography{bibfile}
\end{document}